\documentclass{article}

\usepackage[preprint]{neurips_2025}

\usepackage[utf8]{inputenc} 
\usepackage[T1]{fontenc}    
\usepackage{hyperref}       
\usepackage{url}            
\usepackage{booktabs}       
\usepackage{amsfonts}       
\usepackage{nicefrac}       
\usepackage{microtype}      
\usepackage{xcolor}         
\usepackage{amsmath}
\usepackage{graphicx}
\usepackage{multirow}
\usepackage{colortbl}
\usepackage{caption}
\usepackage{bm}

\title{\textbf{Drive-R1}: Bridging Reasoning and Planning in VLMs for Autonomous Driving with Reinforcement Learning}

\author{
    Yue Li \textsuperscript{1}\thanks{The work was done during Yue Li’s internship at Huawei Noah’s Ark Lab. Email: yueli65@mail.ustc.edu.cn} \quad
    Meng Tian \textsuperscript{2} \quad
    Dechang Zhu \textsuperscript{2} \quad
    Jiangtong Zhu \textsuperscript{2} \quad \\
    \textbf{Zhenyu Lin \textsuperscript{2}} \quad 
    \textbf{Zhiwei Xiong \textsuperscript{1}\thanks{Corresponding Authors. Emails: zwxiong@ustc.edu.cn, zhaoxinhai1@huawei.com}} \quad
    \textbf{Xinhai Zhao \textsuperscript{2}\footnotemark[2]}
    \\
    \textsuperscript{1} University of Science and Technology of China \\
    \textsuperscript{2} Huawei Noah’s Ark Lab \\
}

\begin{document}

\maketitle

\begin{abstract}
  Large vision-language models (VLMs) for autonomous driving (AD) are evolving beyond perception and cognition tasks toward motion planning. However, we identify two critical challenges in this direction: (1) VLMs tend to learn shortcuts by relying heavily on history input information, achieving seemingly strong planning results without genuinely understanding the visual inputs; and (2) the chain-of-thought (COT) reasoning processes are always misaligned with the motion planning outcomes, and how to effectively leverage the complex reasoning capability to enhance planning remains largely underexplored. In this paper, we start from a small-scale domain-specific VLM and propose \textbf{Drive-R1} designed to bridges the scenario reasoning and motion planning for AD. \textbf{Drive-R1} first undergoes the supervised finetuning on a elaborate dataset containing both long and short COT data. \textbf{Drive-R1} is encouraged to reason step-by-step from visual input to final planning decisions. Subsequently, \textbf{Drive-R1} is trained within a reinforcement learning framework that incentivizes the discovery of reasoning paths that are more informative for planning, guided by rewards based on predicted trajectories and meta actions. Experimental evaluations on the nuScenes and DriveLM-nuScenes benchmarks demonstrate that \textbf{Drive-R1} achieves superior performance compared to existing state-of-the-art VLMs. We believe that \textbf{Drive-R1} presents a promising direction for bridging reasoning and planning in AD, offering methodological insights for future research and applications.

\end{abstract}

\section{Introduction}
Autonomous driving (AD) systems aim to enable vehicles to perceive, understand, and interact with their environments in a safe and intelligent manner. Among the core modules in AD pipelines, motion planning plays a central role in determining the future actions, balancing the safety, efficiency, and comfort in real-world driving scenarios. Given observations of the environment and other agents, trajectory prediction directly influence the subsequent low-level control.

Traditional motion planning methods often rely on manually crafted rules~\cite{chen2015deepdriving,sauer2018conditional,fan2018baidu} that operate under simplified assumptions of the environment and agent behaviors. 
While these approaches offer interpretability and robustness in structured scenarios, they typically struggle to handle uncertainty, multi-agent interaction, and diverse traffic patterns. Recently, deep learning-based methods~\cite{hu2022stp3,hu2023planning,jiang2023vad} have shown remarkable success in trajectory prediction by leveraging large-scale driving datasets. 
These methods, comprised of encoder-decoder architectures or spatio-temporal transformers, model the complex agent dynamics and social interactions. The trajectory prediction lacks the interpretability and still faces limitations in reasoning under ambiguous contexts, adapting to open-world conditions and long-tailed events. The emergence of large vision-language models (VLMs) have introduced new opportunities for enhancing AD systems. Recent methods ~\cite{xu2024drivegpt4,nie2025reason2drive,marcu2312lingoqa,mao2023gpt,tian2024drivevlm,sima2023drivelm} have demonstrated promising results in scene perception, description, and decision-making with analysis in open form visual question answers task. Further, the methods ~\cite{mao2023gpt,huang2024making,wang2024omnidrive,jiang2025alphadrive} extend the perception and cognition tasks to motion planning task, with some output interpretable decision processes. 
\begin{figure}[!t]
  \centering
  \includegraphics[width=\textwidth]{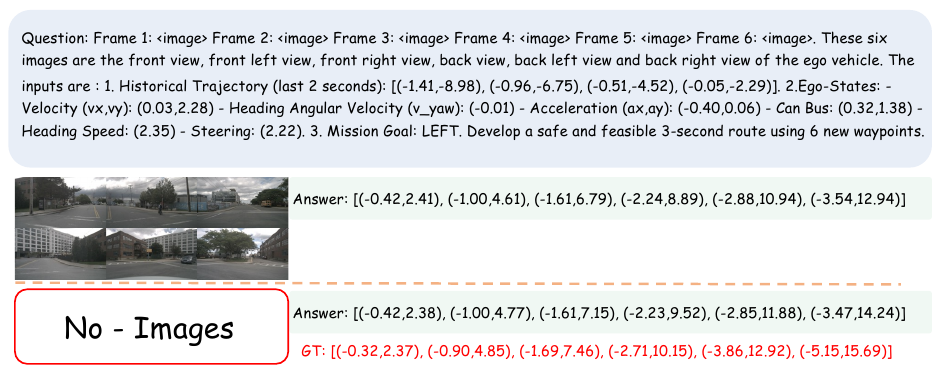}
  \caption{Inference results with and without visual inputs from the model which is trained to predict trajectory without chain of thoughts.}
  \label{fig:woimg}
  \vspace{-0.6cm}
\end{figure}

However, several fundamental limitations remain insufficiently addressed in current VLM-based planning systems. 1) \textbf{The utilization of visual-grounded response in motion planning is limited or even entirely absent.} Recent VLM-based approaches~\cite{tian2024drivevlm,hwang2024emma} achieve strong open-loop metrics by predicting trajectories from image-text inputs, often with short or no chain-of-thought (COT) reasoning. Early GPT-driver~\cite{mao2023gpt} revealed that transforming all perceptual and historical information into textual inputs and using a pure large language model alone can already produce competitive planning performance. To further probe this, we train a general VLM to predict trajectories without COT supervision. At the test time, we ablate the visual input entirely and find that the model performs comparably or even better than with full multi-model inputs. 
The observation indicates that VLMs for motion planning under-utilize the visual modality and heavily rely on textual priors, especially historical motion cues, raising concerns about their visual grounding and generalization.
2) \textbf{The COT reasoning traces are always misaligned with the motion planning outcomes.} Leveraging the reasoning capability to enhance planning performance remains largely underexplored. Recent methods ~\cite{tian2024drivevlm,huang2024making} engages in sequential question-answering to arrive at the final trajectory prediction.
While such methods introduce interpretable intermediate steps, the reasoning remains loosely coupled with planning. We further observe that even when a domain-specific (DS) driving model is trained on motion planning datasets with CoT reasoning, it often falls into a reasoning trap. First, the reasoning patterns learned from CoT data especially designed for complex scenarios may introduce unnecessary analysis in simple cases, leading to overthinking and ultimately injecting noise into the planning output. Second, even manually annotated CoT cannot guarantee precise alignment with the ground-truth trajectories, as natural language reasoning tends to be coarse-grained and ambiguous compared to the fine-grained numerical representation required for planning~\cite{jiang2025alphadrive}.


To address the aforementioned challenges and bridge the gap between scenario reasoning and trajectory-level motion planning in AD, we introduce \textbf{Drive-R1} tailored for vision-language reasoning and trajectory prediction. We begin with a general VLM, InternVL2~\cite{chen2024internvl}, and adapt it to the AD domain by post-training on a large-scale, self-collected dataset comprising 3 million samples. This DS model is endowed with strong perception and scene understanding capabilities, forming a foundation for downstream planning tasks. 

To enable reasoning-aware planning, we construct a structured annotation pipeline that generates CoT data according to key domains in real-world AD~\cite{li2025fine}, including traffic knowledge understanding, general element recognition, traffic graph generation, target attribute comprehension, and ego decision-making and planning. The resulting CoT dataset contains approximately 4,000 samples, categorized into short and long CoT based on the complexity of the driving scenarios: short CoT correspond to relatively simple situations that require minimal deliberation, whereas long CoT are designed for complex, multi-agent, or rule-intensive scenes demanding richer step-by-step reasoning. During the supervised learning stage, \textbf{Drive-R1} is trained on the elaborate dataset to learn to reason from visual observations toward final planning outputs in an interpretable and structured manner. This stage is crucial for encouraging grounded reasoning and mitigating the tendency to overfit to historical trajectory patterns or exploit dataset shortcuts. 

To further align the textual reasoning and numerical trajectory planning, we introduce the reinforcement learning (RL) inspired by the success of recent RL approaches~\cite{guo2025deepseek,huang2025vision}. Specifically, \textbf{Drive-R1} employs the Group Relative Policy Optimization (GRPO), which performa optimization over a set of candidate solutions. The relative optimization mechanism is particularly suitable for motion planning, where multiple plausible trajectories may exist under the same driving scenario. By leveraging comparisons across diverse candidates, GRPO encourages the model to discover reasoning paths that generalize well across variations, rather than overfitting to a single deterministic trajectory, thereby enhancing both planning robustness and generalization. The reward design in GRPO integrates four components: trajectory accuracy, meta-action correctness, repetition penalty, and output format compliance. Among them, the trajectory reward captures outcome-level planning quality, while the meta-action reward reflects the reasoning process quality. These two reward signals are complementary, further promoting effective alignment between reasoning and planning within the \textbf{Drive-R1} framework.

We conduct extensive experiments on both the nuScenes~\cite{caesar2020nuscenes} dataset and the DriveLM-nuScenes~\cite{sima2023drivelm} dataset. Our proposed \textbf{Drive-R1} achieves state-of-the-art performance on the trajectory prediction task, demonstrating its effectiveness in visual-grounded motion planning. Furthermore, we perform comprehensive ablation studies on DriveLM-nuScenes, investigating the impact of various components, including the GRPO on model in different phases, the number of rollouts, and the influence of different reward functions. Our contribution can be summarized as follows:
\begin{itemize}
    \item We identify two key challenges in applying VLMs to motion planning: (i) the over-reliance on historical textual inputs leads to shortcut learning, weakening the visual grounding; and (ii) the misalignment between reasoning chains and planning outputs hinders effective integration of interpretability and decision quality.
    \item We propose \textbf{Drive-R1}, a DS VLM tailored for AD, which connects visual-grounded reasoning to trajectory planning. Our approach incorporates supervised learning on a carefully constructed dataset containing both long and short CoT annotations, followed by RL with GRPO to further align reasoning quality with planning performance.
    \item We conduct extensive experiments on nuScenes and DriveLM-nuScenes, where \textbf{Drive-R1} achieves state-of-the-art results on trajectory prediction.
\end{itemize}

While our work represents a straightforward exploration of integrating VLM into the motion planning pipeline, the insights gained from \textbf{Drive-R1} may offer valuable guidance for future efforts toward the practical deployment of VLMs in AD systems.

\section{Related Work}
\subsection{Vision-language Models for Autonomous Driving}
\label{gen_inst}
The integration of VLMs into AD has recently gained significant attention, aiming to unify perception, reasoning, and planning within a single framework. Existing works in this field can be broadly divided into two categories: scene reasoning-oriented models, and planning and control-oriented models. The first focuses on scene understanding and reasoning~\cite{marcu2312lingoqa,ma2023dolphins,sima2023drivelm,nie2025reason2drive,ding2024holistic}, where VLMs are used to analyze visual environments through natural language, often leveraging question-answering or chain-of-thought reasoning to enhance transparency and trustworthiness. Planning and control-oriented models~\cite{tian2024drivevlm,wang2024omnidrive,xu2024drivegpt4,pan2024vlp,chen2025asynchronous,shao2024lmdrive}, on the other hand, aim to directly generate actionable outputs such as trajectories or control signals from visual and linguistic inputs. These systems often leverage large-scale data and unified modeling to perform planning implicitly within the language model, with or without intermediate reasoning steps. In this paper, we focus on trajectory prediction and find that models can achieve competitive planning performance even with limited or no visual input, suggesting a potential over-reliance on linguistic or historical features and insufficient grounding in visual observations.



\begin{figure}[!t]
  \centering
  \includegraphics[width=\textwidth]{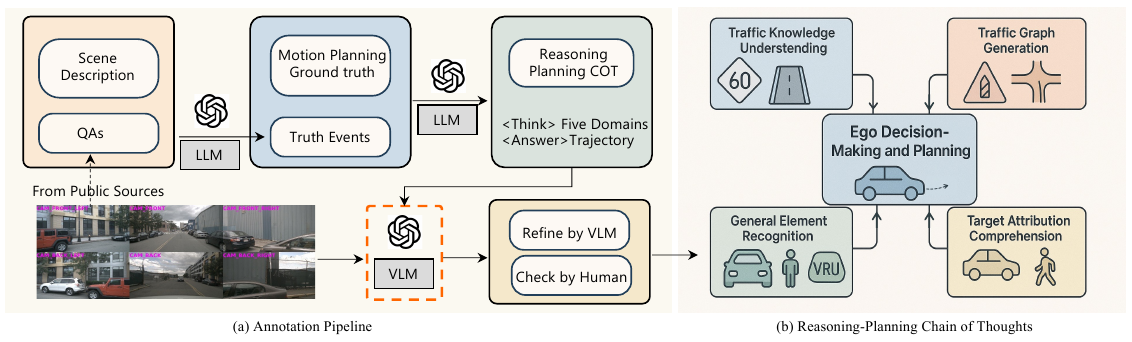}
  \caption{The RP-COT data annotation pipeline.}
  \label{fig:anno_pipe}
\end{figure}

\subsection{Reinforcement Learning}
RL has played a pivotal role in the recent evolution of large language models, particularly in aligning model outputs with human preferences or task-specific objectives. Early developments, such as proximal policy optimization~\cite{schulman2017proximal} and direct policy optimization~\cite{rafailov2023direct}
have been widely adopted in general-purpose LLMs to improve response helpfulness and safety from human feedback, demonstrating that large models could benefit from post-training optimization beyond supervised learning, enabling them to reason and act in more aligned and consistent ways. Recent
Group Relative Policy Optimization (GRPO)~\cite{guo2025deepseek} proposes a group-wise relative optimization strategy, which compares the relative merits of multiple output candidates instead of optimizing based solely on absolute reward values. GRPO has shown strong potential in complex reasoning tasks by encouraging models to explore interpretable thought processes rather than shortcutting to answers. Building on this, RL has been extended to VLMs to enhance their ability to perform multi-step reasoning grounded in visual inputs ~\cite{huang2025vision,jiang2025alphadrive,lai2025med}. AlhpaDrive~\cite{jiang2025alphadrive} introduced RL to high-level planning and ReCogDrive~\cite{li2025recogdrive} claimed the gap between the discrete language space and the continuous action space. Aligning textual reasoning with numerical outputs like trajectories in AD presents unique challenges, requiring designs that balance process-level and result-level precision. In this paper, our work 
still pursues the alignment between reasoning and planning in the discrete space.

\section{\textbf{Drive-R1}}
\label{headings}
\textbf{Drive-R1} aims to bridge scenario-level reasoning and trajectory planning for AD through a combination of SFT and RL. We begin by introducing the construction of the Reasoning-Planning chain of thought (RP-CoT) dataset, which encodes intermediate reasoning steps aligned with planning outcomes. 
Then  we detail the supervised training phase, highlighting the initial capabilities the model must acquire to address the challenges discussed above. Finally, we describe the RL procedure, which leverages carefully designed reward functions to further align textual reasoning with numerical trajectory prediction, enhancing both interpretability and planning performance.

\subsection{RP-COT Data Annotation}
Following the five key domains identified in~\cite{li2025fine} as fundamental to motion planning, i.e., traffic knowledge understanding, general element recognition, traffic graph generation, target attribute comprehension, and ego decision-making and planning, we construct the RP-CoT dataset. RP-CoT is designed to bridge high-level scenario reasoning with low-level trajectory outputs in AD. The scenes are selected from nuScenes~\cite{caesar2020nuscenes}. Each annotation sample in RP-CoT includes step-by-step textual reasoning that reflects a structured understanding of the driving scene, ultimately grounded in a precise trajectory decision. 

As shown in Fig.\ref{fig:anno_pipe}, the annotation pipeline is semi-automatic. We begin by collecting driving scenes from publicly available sources~\cite{sima2023drivelm,qian2024nuscenes,inoue2024nuscenes}, which are annotated with scene descriptions and question-answer (QA) pairs. According to the hundreds of QAs of the scene, ChatGPT first generates the truth events, which are structured representations of the underlying reasoning rationale. Next, based on the ground-truth events and the motion planning information (history trajectory, ego status, meta action), ChatGPT generates RP-COT data through the five domains. Each sample includes the <think></think> section that explains reasoning steps and the <trajectory></trajectory> section that specifies the future trajectory (6 points with 3 seconds). To ensure the generated RP-CoTs are grounded in visual reality, the VLM, GPT-4o, is employed to refine these outputs by aligning them with scene content. Finally, all annotations are checked by human annotators to guarantee consistency, correctness, and planning validity.

Our annotation pipeline systematically decomposes the visual-linguistic information into reasoning stages aligned with the aforementioned domains. This structured format enables the model to learn interpretable reasoning paths that progressively lead to planning actions, laying a strong foundation for subsequent learning stages.

\setlength{\tabcolsep}{4.8pt}
\begin{table*}[!t]
\caption{The preliminary experimental results validated on 799 samples from DriveLM-nuScenes~\cite{malla2023drama}. BA and DS are the base and domain-specific models. WI denotes inference without image inputs.}
\centering
\label{tab:ablasft}
\begin{tabular}{c|cc|cc|cccc|cccc}
\hline
\multirow{2}{*}{Models} & \multicolumn{2}{c|}{RP-COT} & \multicolumn{2}{c|}{Trainning Phase} & \multicolumn{4}{c|}{L2(m)$\downarrow$} & \multicolumn{4}{c}{Collision$\downarrow$} \\ \cline{2-13} 
                 & Long         & Short       & SFT         & RFT         & 1s   & 2s   & 3s   & Avg  & 1s    & 2s    & 3s    & Avg   \\
\hline
BA         & $\times$    & \checkmark         & \checkmark           & $\times$           & 0.25 & 0.55 & 0.97 & 0.59 & 0.00  & 0.09  & 0.56  & 0.22  \\
\rowcolor[HTML]{EFEFEF}
BA-WI        & $\times$    & \checkmark         & \checkmark           & $\times$           & 0.25 & 0.53 & 0.90 & 0.56 & 0.00  & 0.03  & 0.48  & 0.18  \\
\hline
DS         & $\times$    & \checkmark         & \checkmark           & $\times$           & 0.18 & 0.42 & 0.76 & 0.45 & 0.00  & 0.03  & 0.46  & 0.16  \\
DS         & \checkmark     & $\times$        & \checkmark           & $\times$           & 0.24 & 0.53 & 0.91 & 0.56 & 0.00  & 0.19  & 0.61  & 0.27  \\
DS            & \checkmark & \checkmark          & \checkmark           & $\times$           & 0.19 & 0.39 & 0.67 & 0.41 & 0.00  & 0.03  & 0.29  & 0.11  \\
\hline
BA            &-&-          & $\times$           & \checkmark           & 0.37 & 0.75 & 1.22 & 0.78 & 0.00  & 0.16  & 0.84  & 0.33  \\
DS            &-&-          & $\times$           & \checkmark           & 0.26 & 0.55 & 0.93 & 0.58 & 0.00  & 0.19  & 0.61  & 0.27  \\
DS        & \checkmark & \checkmark              & \checkmark           & \checkmark           & 0.17 & 0.35 & 0.60 & 0.37 & 0.00  & 0.00  & 0.30  & 0.10 \\
\hline
\end{tabular}
\vspace{-0.2cm}
\end{table*}

\begin{figure}[!t]
  \centering
  \includegraphics[width=\textwidth]{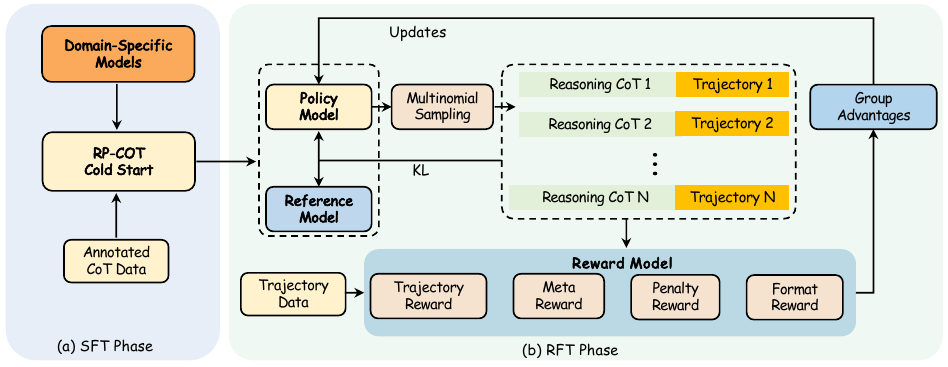}
  \caption{The overview of the proposed \textbf{Drive-R1}, which comprises the supervised fine-tuning (SFT) and reinforcement fine-tuning (RFT) phases.}
  \label{fig:pipeline}
  \vspace{-0.6cm}
\end{figure}

\subsection{Supervised Fine-tuning Phase}
As discussed above, the utilization of visual-grounded response in motion planning is limited or even absent. As shown in Table~\ref{tab:ablasft}, directly training a general-purpose VLM (Internvl2-4B) to output trajectories without CoT supervision can yield surprisingly competitive performance. However, we observe a counterintuitive outcome: the model performs better when visual inputs are ignored, indicating a strong reliance on historical textual context over visual perception.
We attribute this phenomenon to two key factors: (1) the model lacks sufficient familiarity with DS tasks in AD and (2) the model are more sensitive to historical motion cues than to scene-level visual information. To address this, we first perform full-parameter finetuning of an InternVL2-4B model on a large-scale, DS dataset comprising 3 million AD QAs, which are collected from public sources~\cite{parikh2024idd,lu2024can,marcu2312lingoqa,sima2023drivelm,cao2024maplm,li2022coda,malla2023drama,kim2019CVPRhad,ma2023dolphins,wang2023openlane,guo2023visualrsk10k,xu2024drivegpt4,mao2023gpt}. The DS model from the first SFT stage significantly mitigates the overreliance on historical information and enhances its general understanding of AD scenarios. Nevertheless, the gap between visual-informed and vision-agnostic reasoning remains narrow. Further, we incorporate the previously constructed RP-CoT dataset into the second SFT stage. Through supervised CoT supervision, the model is encouraged to form visual-grounded reasoning paths across key domains, solving the dependency on textual history information and thereby promoting more robust, perception-aware planning behavior.

On the other hand, the COT reasoning traces are always misaligned with the motion planning outcomes. The experimental results in Table~\ref{tab:ablasft} show that applying long CoT supervision during the SFT stage lead to a decline in performance compared to directly supervising the final trajectory output. Interestingly, similar observations occur in other domains. Recent researches~\cite{wang2024enhancing, tan2025reason} report that for tasks involving spatial reasoning or numerical sensitivity, models trained with CoT supervision often underperform compared to those trained with direct answer supervision. We hypothesize that the observed performance degradation may stem from two primary factors: (1) the limited representation capacity of small-scale models, which restricts their ability to accurately encode and utilize complex reasoning paths~\cite{li2025small} and (2) the differing tolerance to errors of the models between textual and numerical outputs. Specifically, reasoning texts generated during CoT supervision may contain semantic inconsistencies or hallucinations, either due to imperfect annotation quality or intrinsic limitations of the model. While such errors may have negligible impact on the interpretability or plausibility of the textual reasoning itself, they can propagate to the numerical prediction stage, e.g., trajectory prediction, where small deviations are amplified into significant planning errors.

To mitigate the negative impact of indiscriminate CoT supervision, we introduce a fast-and-slow thinking strategy in the second SFT stage. The core idea is to adapt the complexity of reasoning supervision to the difficulty of each driving scenario. Specifically, we categorize CoT supervision into short CoT and long CoT, depending on the reasoning demand: short CoT corresponds to relatively simple scenarios requiring minimal deliberation, while long CoT is designed for complex, multi-agent, or rule-intensive scenes that demand richer step-by-step reasoning. We begin by training a model to directly generate trajectory outputs without CoT supervision. This specific model is then used to assess the reasoning complexity of each scene, serving as a proxy for scenario difficulty. Scenes with low planning metrics are assigned short CoT supervision, while those with high planning metrics are paired with long CoT annotations. As shown in Table~\ref{tab:ablasft}, models fine-tuned with this adaptive fast-and-slow thinking strategy achieve the best overall performance, validating its effectiveness in balancing long and short COT.



\subsection{Reinforcement Learning Phase}
DeepSeek-R1~\cite{guo2025deepseek} demonstrates that RL frameworks like GRPO can effectively elicit long CoT reasoning abilities of large language models. However, subsequent studies~\cite{yue2025does,chu2025sft} have shown that the reasoning paths produced by RL-finetuned models already exist with high probability in the output distribution of base model, i.e., problems solvable by the RL model can also be addressed by the base model through sufficient sampling. Building upon these insights, we adopt GRPO not as a means to unlock fundamentally new capabilities, but rather as a post-training alignment mechanism to improve the efficiency and consistency for further aligning the reasoning and planning.

\subsubsection{Algorithm}
Specifically, for each question q, GRPO~\cite{guo2025deepseek} samples a group of candidate outputs $
\{o_1, o_2, \cdots, o_G\}$ from the old policy $\pi_{\theta_{\text{old}}}$, and subsequently updates the current policy $\pi_\theta$ by maximizing the following objective:

\begin{align}
\mathcal{J}_{\text{GRPO}}(\theta) &= \mathbb{E}_{q, \{o_i\}_{i=1}^{G} \sim \pi_{\theta_{\text{old}}}(O|q)} \left[ \frac{1}{G} \sum_{i=1}^{G} w_iA_i - \beta \mathbb{D}_{\text{KL}}(\pi_\theta \| \pi_{\text{ref}}) ) \right]
\end{align}

\begin{align}
w_i = \min ( \frac{\pi_\theta(o_i|q)}{\pi_{\theta_{\text{old}}}(o_i|q)}, \text{clip}( \frac{\pi_\theta(o_i|q)}{\pi_{\theta_{\text{old}}}(o_i|q)}, 1 - \epsilon, 1 + \epsilon ), \quad  A_i = \frac{r_i - \text{mean}(\{r_1, r_2, \dots, r_G\})}{\text{std}(\{r_1, r_2, \dots, r_G\})}
\end{align}

where the KL loss is calculated by $\mathbb{D}_{\text{KL}}(\pi_\theta \| \pi_{\text{ref}}) = \frac{\pi_{\text{ref}}(o_i|q)}{\pi_\theta(o_i|q)} - \log \frac{\pi_{\text{ref}}(o_i|q)}{\pi_\theta(o_i|q)} - 1$, $
\{r_1, r_2, \cdots, r_G\}$ are the rewards of the candidate outputs.

\subsubsection{Rewards}
To better align the intermediate reasoning steps with final motion planning outcomes, we design a composite reward function that balances both process-level and outcome-level results. The total rewards comprise the following components:

\noindent
\textbf{Trajectory Reward} measures the accuracy of the predicted trajectory by computing the L2 distance between the predicted trajectory $\hat{\tau}$ and the ground truth $\tau$. The raw distance is then mapped using a sigmoid-based transformation: $R_{\text{traj}} = \frac{2e^{-d}}{1 + e^{-d}}$, $d = \lVert \hat{\tau} - \tau \rVert_2$.

\noindent
\textbf{Meta-Action Reward} assesses the high-level planning decisions in the reasoning section, including the short-term lateral and longitudinal decisions. Each contributing 0.5 to the total reward score.

\noindent
\textbf{Repetition Penalty} penalizes the generation of redundant or repetitive reasoning steps within the CoT to encourage concise and efficient planning rationale~\cite{yeo2025demystifying}.

\noindent
\textbf{Format Reward} ensures structural correctness of the output format.

Importantly, our analysis shows that the result-oriented trajectory reward and the process -oriented meta-action reward are positively correlated. 

\subsubsection{Training}
Through extensive experiments, we observe that effective RL in the context of motion planning is highly dependent on the model’s prior alignment with the AD domain. When applied to models without sufficient domain adaptation, reinforcement signals often result in unstable updates or limited policy improvement, suggesting that the capacity to interpret structured driving scenarios is a prerequisite for successful policy refinement. Consequently, we perform RL on a model that has been supervised via two-stage fine-tuning, as introduced in the SFT phase. Building on such warm-up, RL further amplifies the synergy between visual-grounded reasoning and motion planning, leading to the most significant performance gains observed in our experiments.

\section{Experiments}
\label{others}
\subsection{Datasets and Baselines}
In the first SFT phase, the domain specific data are collected from a diverse set of AD datasets, including~\cite{parikh2024idd,lu2024can,marcu2312lingoqa,sima2023drivelm,cao2024maplm,li2022coda,malla2023drama,kim2019CVPRhad,ma2023dolphins,wang2023openlane,guo2023visualrsk10k,xu2024drivegpt4,mao2023gpt} with 3 million samples. The QAs are built following the five key domains, and include the single-view, multi-view, and sequential image inputs. In the second SFT phase, RP-COT dataset are construct from the annotations in ~\cite{sima2023drivelm,qian2024nuscenes,inoue2024nuscenes} with the number of samples 4,072.
When compared on the 6019 validation samples on nuScenes~\cite{caesar2020nuscenes}, the numbers of long and short RP-COT are 24058 and 4072. When compared on the 799 validation samples on DriveLM-nuScenes~\cite{sima2023drivelm}, the numbers of long and short RP-COT are 2036 and 2036. In the RL phase, the samples are selected from those in 4072 RP-COT datasets.

We benchmark \textbf{Drive-R1} against both end-to-end and vision-language planning baselines. The former includes ST-P3~\cite{hu2022stp3},UniAD~\cite{hu2023planning} and their modified versions augmented with ego-status inputs~\cite{jiang2023vad}. The latter set of baselines includes DriveVLM~\cite{tian2024drivevlm}, RDA-Driver~\cite{huang2024making}, OmniDrive~\cite{wang2024omnidrive}, and EMMA~\cite{hwang2024emma}. Notably, prior approaches typically output direct trajectory predictions either without reasoning or with short CoT supervision. In contrast, \textbf{Drive-R1} produces both reasoning chains and trajectory outcomes in a unified manner, enabling interpretable and context-sensitive planning.

\subsection{Implementation and Metrics}
The SFT training is conducted based on the official codebase of InternVL2~\cite{chen2024internvl}. The first-stage SFT is trained on 32 V100 nodes with a batch size of 256, while the second-stage SFT is trained on 16 V100 nodes with a batch size of 128. The RL phase is implemented using the ms-swift framework~\cite{zhao2025swift} and trained on 2 V100 nodes with a batchsize of 16 and a rollout of 6. The context length is set to 4096. For evaluation, we adopt the L2 distance and collision rate metrics, following ST-P3~\cite{hu2022stp3}.

\subsection{Results}
Table~\ref{all_result} presents a comprehensive comparison between \textbf{Drive-R1} and existing representative baselines. \textbf{Drive-R1} achieves the lowest average L2 error of 0.31, marginally outperforming EMMA (0.32) ~\cite{hwang2024emma} and OmniDrive (0.33)~\cite{wang2024omnidrive}, while also exhibiting the lowest average collision rate of 0.09. In contrast, several end-to-end methods demonstrate competitive L2 metrics, yet suffer from relatively higher collision rates. This suggests that while these models may fit the trajectory well numerically, they may lack robustness in safety-critical aspects of planning. Among VLM-based baselines, \textbf{Drive-R1} consistently achieves better planning quality and safety. Notably, compared with RDA-Driver~\cite{huang2024making} and OmniDrive~\cite{wang2024omnidrive}, our model demonstrates both improved trajectory precision and reduced collision risks, validating the effectiveness of reasoning-aligned trajectory generation.

\setlength{\tabcolsep}{10pt}
\begin{table}[!t]
\caption{Overall comparison with baselines on the nuScenes~\cite{caesar2020nuscenes} validation.}
\label{all_result}
\centering
\begin{tabular}{c|cccc|cccc}
\hline
\multirow{2}{*}{Models} & \multicolumn{4}{c|}{L2(m)$\downarrow$} & \multicolumn{4}{c}{Collision$\downarrow$} \\ \cline{2-9}
                        & 1s   & 2s   & 3s   & Avg  & 1s    & 2s    & 3s    & Avg   \\ \hline
ST-P3~\cite{hu2022stp3}                  & 1.33 & 2.11 & 2.90 & 2.11 & 0.23  & 0.62  & 1.27  & 0.71  \\
UniAD~\cite{hu2023planning}& 0.48 & 0.96 & 1.65 & 1.03 & 0.05  & 0.17  & 0.71  & 0.31  \\
UniAD-E~\cite{hu2023planning}                 & 0.20 & 0.42 & 0.75 & 0.46 & 0.02  & 0.25  & 0.84  & 0.37  \\
VAD-E~\cite{jiang2023vad}                   & 0.17 & 0.34 & 0.60 & 0.37 & 0.07  & 0.10  & 0.24  & 0.14  \\
DriveVLM~\cite{tian2024drivevlm}                & 0.18 & 0.34 & 0.68 & 0.40 & 0.10  & 0.22  & 0.45  & 0.27  \\
RDA-Driver~\cite{huang2024making}              & 0.17 & 0.37 & 0.69 & 0.40 & 0.01  & 0.05  & 0.26  & 0.10  \\
OmniDrive~\cite{wang2024omnidrive}              & 0.14 & 0.29 & 0.55 & 0.33 & 0.00  & 0.13  & 0.78  & 0.30  \\
EMMA~\cite{hwang2024emma}                    & 0.14 & 0.29 & 0.54 & 0.32 & -     & -     & -     & -     \\
\textbf{Drive-R1} (Ours)               & 0.14 & 0.28 & 0.50 & 0.31 & 0.02  & 0.06  & 0.19  & 0.09 \\  \hline
\end{tabular}
\end{table}

\begin{figure}[!t]
  \centering
  \includegraphics[width=\textwidth]{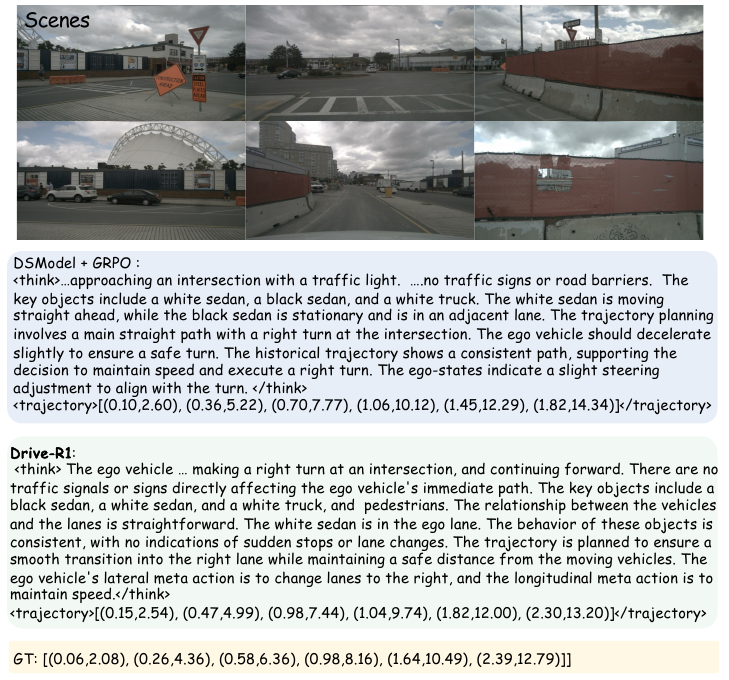}
  \caption{An inference results from different models.}
  \vspace{-0.3cm}
\end{figure}

\subsection{Ablation Studies}
We conduct extensive ablation experiments on the DriveLM-nuScenes~\cite{sima2023drivelm} to investigate the effects of RP-COT input types and RL configurations.

\noindent
\textbf{Effect of COT Length in SFT Stage.}
As shown in Table ~\ref{tab:ablasft}, we evaluate the influence of long and short RP-COTs in the second SFT stage. Models trained with only short RP-COTs or long RP-COTs underperform those trained with both long and short RP-COTs, suggesting that applying a uniform CoT strategy across diverse scenarios is suboptimal. Instead, combining both short and long CoTs better equips the model to handle a wider variety of AD contexts, leveraging both concise and elaborate reasoning chains.

\noindent
\textbf{Effectiveness of RL on Different Model Bases.}
In Table ~\ref{tab:ablasft}, we further assess how RL impacts different model variants. The DS model, trained on the first SFT stage, benefits from RL than the base model. Further, when incorporating long and short RP-COT pattern, the model significantly benefit from the RL phase, reducing both trajectory deviation and collision rates. This underscores the necessity of prior domain alignment before performing RL fine-tuning.
\setlength{\tabcolsep}{6pt}
\begin{table*}[!t]
\centering
\caption{Ablation Studies of the reward designs and the number of rollouts in GRPO. T., F., R., M. represent the trajectory reward, format reward, repetition penalty, and meta-action reward.}
\label{tab:abla}
\begin{tabular}{ccc|c|cccc|cccc}
\hline
\multicolumn{3}{c|}{Rewards}                    & Rollouts & \multicolumn{4}{c|}{L2(m)$\downarrow$} & \multicolumn{4}{c}{Collision$\downarrow$} \\ \hline
T.\&F. & R. & M. & Nums       & 1s   & 2s   & 3s   & Avg  & 1s    & 2s    & 3s    & Avg   \\
\checkmark          & $\times$          & $\times$           & 6       & 0.17 & 0.35 & 0.60 & 0.37 & 0.06  & 0.06  & 0.42  & 0.18  \\
\checkmark               & \checkmark          & $\times$           & 6       & 0.17 & 0.34 & 0.59 & 0.37 & 0.06  & 0.06  & 0.30  & 0.14  \\
\checkmark               & \checkmark          & \checkmark           & 6       & 0.17 & 0.35 & 0.60 & 0.37 & 0.00  & 0.00  & 0.30  & 0.10  \\
\hline
\checkmark               & \checkmark          & \checkmark           & 12      & 0.17 & 0.35 & 0.60 & 0.37 & 0.00  & 0.03  & 0.25  & 0.11  \\
\checkmark               & \checkmark          & \checkmark           & 24      & 0.17 & 0.35 & 0.60 & 0.37 & 0.00  & 0.03  & 0.21  & 0.08 \\
\hline
\end{tabular}
\end{table*}

\noindent
\textbf{Impact of Reward Design and Rollout Numbers.}
In Table ~\ref{tab:abla}, we evaluate how various reward components and rollout counts affect model performance. The inclusion of meta-action rewards and repetition penalties leads to consistent improvements in collision rates (e.g., 0.14 to 0.10), highlighting their effectiveness in guiding safer planning behavior. However, for models with relatively small capacity, simply increasing the number of rollouts does not always yield stable or consistent performance gains. For instance, although the collision rate decreases from 0.10 to 0.11 and to 0.08 when the number of rollouts increases from 6 to 12 and to 24, we observe that the training becomes unstable beyond a certain number of steps under the 24-rollout setting. It is worth noting that the reported result at 24-rollout is extracted before the onset of training collapse.

\section{Conclusion}
In this work, we present \textbf{Drive-R1}, which bridges the structured chain-of-thought reasoning and the trajectory-level motion planning. To address the insufficient visual grounding and the misalignment between reasoning traces and planning outputs observed in existing VLM-based approaches, we construct a domain-specific VLM and augment it with a systematically annotated CoT dataset spanning five essential reasoning domains. Furthermore, we incorporate a RL phase based on GRPO to optimize planning quality for aligning the reasoning process with trajectory outcomes. Comprehensive experiments conducted on nuScenes and DriveLM-nuScenes benchmarks validate the effectiveness of our proposed method. Drive-R1 achieves state-of-the-art performance on trajectory prediction tasks while offering interpretable and structured reasoning capabilities. \textbf{Drive-R1} represents an early exploration toward bridging high-level cognitive reasoning and low-level trajectory planning in AD.  In addition, we conduct extensive experiments on large-scale in-house datasets using Ascend 910 hardware platforms, which further verify the generalizability and robustness of the Drive-R1 framework. We believe that the insights gained from \textbf{Drive-R1} may offer valuable guidance for future efforts toward the practical deployment of VLMs in AD systems.

{
\small
\bibliographystyle{unsrt}
\bibliography{ref}
}

\end{document}